# Lean Methodology for Garment Modernization

[1]Ray Wai Man Kong

[1]Adjunct Professor, City University of Hong Kong, Hong Kong
[1]Modernization Director, Eagle Nice（International） Holding Ltd, Hong Kong

[2]Theodore Ho Tin Kong
[1]Engineer, Master of Science in Aeronautical Engineering, Hong Kong University of Science and Technology, Hong Kong

[3]Tianxu Huang
[3]Senior experimentalist, College of Information Engineering, Guangxi City Vocational University, Chongzuo, 532200, China

[1]Dr.RayKong@cityu.edu.hk, [2]theodorekong@ieee.org, [3]422423782@qq.com

*Abstract*—This article presents the lean methodology for modernizing garment manufacturing, focusing on lean thinking, lean practices, automation development, VSM, and CRP, and how to integrate them effectively. While isolated automation of specific operations can improve efficiency and reduce cycle time, it does not necessarily enhance overall garment output and efficiency. To achieve these broader improvements, it is essential to consider the entire production line and process using VSM and CRP to optimize production and center balance. This approach can increase efficiency, and reduce manufacturing costs, labor time, and lead time, ultimately adding value to the company and factory.

## I. INTRODUCTION

The garment industry is well-known for its complex manufacturing processes and heavy reliance on manual labour and precision. Global garment manufacturers face tough price and cost competition, high labour costs, quality issues, and the need for sustainable environmental practices. These challenges drive companies to adopt manufacturing technology advancements, automation, and lean practices in garment manufacturing.

In traditional manual garment manufacturing processes, factories handle labour-intensive tasks such as fabric and accessories inspection, fabric cutting, stitching, embroidery, down-filling, buttoning, sewing, ironing, trimming, and packing. While the manual process offers flexibility, it relies on skilled labour to maintain quality and consistency.

To fully benefit from robotic automation in the garment industry, incorporating Lean practices is vital. This involves using modernized facilities, automated machinery, and standardized procedures to minimize the need for different sizes of accessories and fabric pieces. By standardizing components, lean practices reduce setup time and wasted efforts, improving overall garment output.

Advancements in manufacturing technology, such as the development of Vacuum Suction Grabbing Technology for automated grabbing of fabric for garment automated sewing machines as referenced by Prof. Ray Wai Man Kong et al. [1], have made the integration of robotic automation in garment manufacturing increasingly feasible. Robotic automation, combined with lean practices, can enhance productivity, efficiency, and quality while reducing labour costs.

However, the relationship between lean practice and automation development in garment modernization has not received much attention in industrial engineering and management studies. Many automated machines have been implemented without considering lean practice, leading to redundant processes and inefficiencies. In addition, a manual step in the garment sewing process upgrades the automated machine instead of manual work, but the single operation in the garment factory is not awarded to calculate the requested number of machines because of the fast changeover of style in the production floor in the high mixed low volume. The singular automated machine is expected to increase the whole garment finished goods output and efficiency. After the singular automated machinery implementation has been done, the result has not achieved the expected efficiency improvement. One of the reasons is the garment factory did not implement any lead theories in the lead production, did not calculate the capacity requirement plan and did not know which operation was a bottleneck of the process. At the beginning stage, there is a lack of methodology in lean modernization. This article aims to address the knowledge gap and provide insights into the successful implementation of lean practices and automation in garment manufacturing.

## II. BACKGROUND

*2.1 Garment Manufacturing Background*

Referring to Sheikh Shohanur Rahman, Abdul Baten, Manjurul Hoque, Md. Iqbal Mahmud [2], the garment industry is a traditional industry that faces worldwide competition. Designing/creating clothing patterns, cutting, sewing, and packaging are the four major elements of the garment production process. The sewing step is the most challenging phase since it requires the completion of many activities. The sewing line is equipped with a series of workstations, each of which processes a distinct operation following a predetermined order. It is observed that the condition of the sewing floor is in a haphazard situation. There are so many garment part stocks between the operators and the output is quite less than the input. Precise work distribution is not maintained by the operators. Materials are used to travel long distances. The quality is not controlled by the line supervisors as they are not strict enough. Non-value-added time is increased because the total completion time is delayed by the workers' reworks. So, a smooth streamlined, and continuous process flow is required to avoid all such unwanted occurrences.

*2.2 Isolated Automated Machinery Apply in Garment Manufacturing*

The use of automated machinery alone is not enough to drive comprehensive process improvement, mainly due to the lack of Value Stream Mapping (VSM) within the framework of Lean Practice (LP). While specific automated machines have enhanced productivity by reducing cycle time in the previous operation, the next operation is unable to increase production capacity to handle the work in progress from the upstream operation. Automating the upstream process has led to a backlog of garment parts in the next operation, as there has been no consideration for improving the overall production capacity of the process. As a result, the overall production output at the end operation cannot be increased or productivity improved as if the isolated automated machine has been applied to only part of the process. The use of VSM provides the advantage of identifying extended process time, queue time, and idle time, enabling the optimization of the production process. Lean practices offer tools such as just-in-time (JIT), VSM, and pull-in systems to help the production line identify any bottlenecks in the entire production process.

*2.3 Insufficient Capacity Requirement Plan in the Whole Garment Manufacturing Process*

Capacity requirement planning (CRP) is an effective tool for calculating machine capacity demand and supply during digital modernization. The required machine capacity is determined based on the machine lead time per style from the shop or job order. Each style and size can identify the machine's lead time. The total demand quantity of job orders per workstation and machine can be multiplied by the machine lead time by style and size to find out the total required machine hour and labour hour of the process by machine.

Based on the available number of machines and available time of the machine per day or per shift, the available total machine capacity per process can be calculated per hour, shift, or day. If the total number of required machine hours per machine (called machine demand capacity) is greater than the total available machine hours (called machine supply capacity), it means that the total demand for machine hours is greater than the total supply of machine-hours, which becomes the bottleneck process. The garment workpieces are jammed on the bottleneck process and machine. The production line balance is not improved if there is considered the isolation of automated machine implementation.

A common problem arises when applying automated machines for a specific process, as it does not necessarily improve the entire production line output or productivity due to an imbalance of production line capacity. The implementation of isolated automation focuses only on the narrow and isolated specific machinery of the process without considering the overall review of the whole process machinery capacity or labour capacity.

## III. LITERATURE REVIEW

*3.1 Lean Principles*

In their book "Lean Thinking," J. P. Womack and D. T. Jones [3] outline five principles of Lean, which are Value, Value Stream, Flow, Pull, and Perfection, and provide suggestions for implementing these principles. They assert that transforming a batch-and-wait factory into a Lean factory can lead to a doubling of productivity, as well as a 90% reduction in lead times and inventories.

The concept of Lean is centered around defining value, with customer determination of the value of an activity or production part being paramount. The authors suggest that real value is typically represented by the direct materials and direct labour required to produce the product, which can be used as "target costs" for a product. The value stream, defined as the activities necessary to move the product or service through an organization's foundational activities, is the second principle of Lean.

The Lean concept has its roots in Japanese traditions and is aimed at minimizing waste throughout the manufacturing process. Waste, in this context, encompasses everything that does not add value from the consumer's perspective. Linear practice (LP) is an approach aligned with Lean that advocates the adoption of techniques such as total quality management (TQM), Kanban, Kaizen, and VSM to reduce waste and enhance business performance, as noted by Womack et al. (1990) [3].

*3.2 Lean Production*

In 1988, Johan Krafick from the article of Abu Hamja et al [4] first introduced Lean Production, which was further developed by Womack et al. in 1990. They presented Lean Production as a system based on the Toyota Production System, aiming to create maximum customer value with efficiency by minimizing inventory and waste. To achieve this, the system's variability needs to be minimized, flow efficiency needs to be optimized, and continuous improvement capability needs to be built. Lean has been widely adopted in the manufacturing and service sectors, but successful implementation is still a subject of intensive research due to the challenges of changing organizational culture and routines. The literature on lean is divided into its definitions and advice on implementation.

Lean manufacturing is viewed as a solution to address rising labour costs and the demand for lower prices, shorter lead times, smaller production runs, higher quality, and social compliance in the garment industry. Its goal is to enhance productivity without compromising compliance. Originating from Japanese practices, lean aims to minimize waste throughout the manufacturing process.

Waste includes everything that doesn't add value to the final products or services from the consumer's perspective. Lean production promotes the adoption of techniques like total quality management, Kanban, Kaizen, and Value Stream Mapping (VSM) to reduce waste and enhance business performance. It encompasses a variety of strategies, tools, methods, and approaches, although it lacks a standardized application structure, practices, methods, or approaches. The main aims of lean practices are to identify and eliminate different types of waste in manufacturing processes and to standardize parts and components to facilitate automation.

*3.3 Lean Practice*

The practice of Lean manufacturing with automation aims to increase production line efficiency, reduce standard applied minutes (SAM), maintain workflow, decrease labour motion, reduce defects, and minimize work-in-progress (WIP) inventory.

Referring to Palash Saha and Subrata Talapatra [5], Lean Practice (LP) encompasses a variety of strategies, tools, methods, and approaches that enable these objectives to be achieved through the implementation of these applications. However, Valamede and Akkari (2020) [6] found that there is no common LP application structure, as well as no specific LP implementation practices, methods, or approaches.

Therefore, the primary objectives of lean practice are to identify various types of waste in fabric and accessories inspection, fabric cutting, stitching, embroidery, down-filling, buttoning, sewing, ironing, trimming, and packing to eliminate 7 types of wastage. Standardization of parts and accessories is an essential factor in applying automation. The goal of lean practice in manufacturing with automation is to increase production, improve line efficiency, reduce SAM, maintain workflow in the sewing section, minimize excess labour motion, reduce defects in the sewing line, and minimize WIP inventory.

*3.4 Robust Capacity Requirements Planning*

In their work, Daisuke Morita and Haruhiko Suwa [7] introduced an Exact Method for Robust Capacity Requirements Planning (CRP) aimed at addressing the capacity line balance problem in machinery scheduling. This method focuses on parallel tasks such as stitching operations and seeks to apply automation and modernization solutions to effectively balance capacity lines. The robust CRP method outlined in the article aims to create a stable load plan that can adapt to dynamic changes in a manufacturing environment. By determining the processing periods of operation orders, the method works to minimize the likelihood of resource requirements exceeding the capacity of corresponding resources.

Based on the approach of capacity planning through the linear programming technique, as discussed in the works of Ayhan Yalcinsoy [8] and Ray Wai Man Kong [9][14], linear programming is an effective method for addressing scheduling, capacity loading, production programming, and machinery capacity planning issues in garment manufacturing. It is commonly employed as a mathematical model for capacity requirement planning and solutions, particularly in the context of automation implementation.

## IV. LEAN METHODOLOGY FOR THE DESIGN OF AUTOMATION IN GARMENT MANUFACTURING MODERNIZATION

The methodology outlined in this article serves as a crucial framework for applying lean practices to the automation and modernization of garment manufacturing. It aims to bridge the knowledge gap and offer valuable insights into the successful implementation of lean practices and automation in this industry. According to lean thinking [4], there are five key steps: 1) identifying value, 2) mapping value stream, 3) creating flow, 4) adopting a pull system, and 5) pursuing perfection. The pursuit of perfection involves embracing lean thinking and continuous process improvement, as well as utilizing the Plan-Do-Check-Act (PDCA) problem-solving iterative method to continuously enhance processes and products.

In reference to [3], the Lean Enterprise Institute (LEI), established by James P. Womack and Daniel T. Jones in 1997, is widely regarded as the leading source for lean wisdom, training, and seminars. According to Womack and Jones, there are five key lean principles: value, value stream, flow, pull, and perfection.

In reference to [6], the engineering design process consists of the following steps: (1) defining the problem, (2) conducting background research and specifying requirements, (3) brainstorming solutions, (4) selecting the best solution, (5) developing the solution, (6) building a prototype and testing, and (7) redesigning and communicating the results. Nikstiwari [10] explained that the engineering design process is a series of steps used to find a solution to a problem. These steps involve problem-solving processes such as identifying objectives and constraints, prototyping, testing, and evaluation. While the design process is iterative and follows a predetermined set of steps, some steps may need to be repeated before moving to the next one. This may vary depending on the project but allows for learning from failures and making improvements. The process involves the use of applied science, mathematics, and engineering sciences to achieve a high level of optimization in meeting the objectives. The steps include problem-solving processes such as identifying objectives and constraints, prototyping, testing, and evaluation.

Figure 1 Lean Methodology Diagram to the Garment Manufacturing Modernization

This article refers to the engineering design process, lean thinking, lean practice, capacity requirement planning and lean production with practical success experiments to create the lean methodology for the design of automation in garment manufacturing modernization.

The following step of lean methodology is for the design of automation in garment manufacturing modernization as referred to in Fig 1.

Figure 1 Lean Methodology Diagram to the Garment Manufacturing Modernization

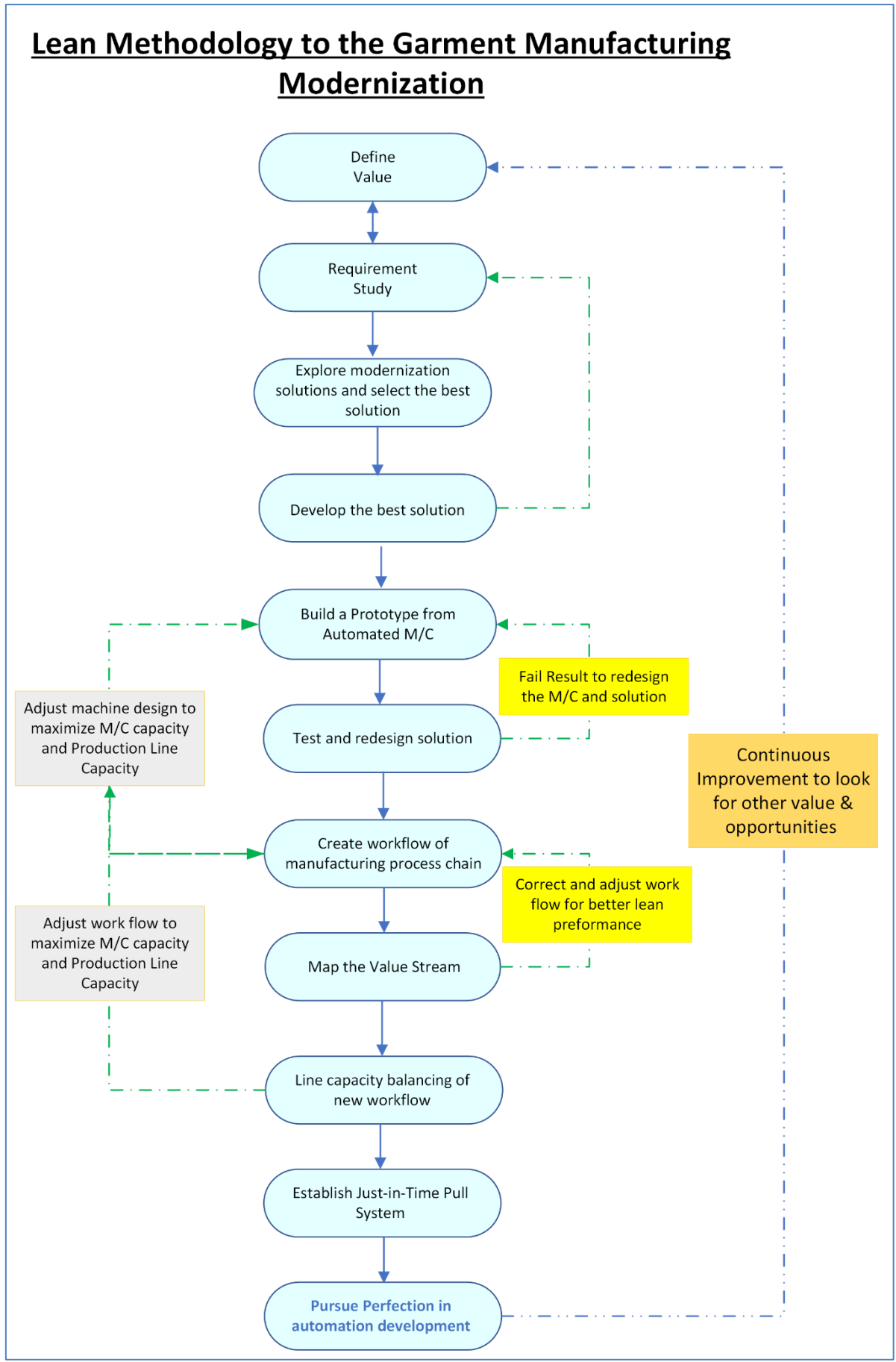

*4.1 Define Value*

Referring to [3], the lean thinking principle underscores the significance of defining value. It is essential to comprehend the concept of value. Value pertains to the purpose for which the designer is creating the automated machine and the lean practice to address a specific problem. The problem represents the area the designer aims to enhance using lean tools and automation. As per the lean principle, value is ascertained using qualitative and quantitative techniques. This principle can be employed to determine which operations and work steps in the process are being eliminated to reduce non-value operations and steps. Before estimating the value at a high level, a requirement study is necessary, involving research on how to apply new automation and advanced manufacturing techniques, and which processes can be improved to provide value, such as shortening the cycle time of standard applied minutes and reducing the human workforce content.

*4.2 Requirement Study*

In the lean methodology for modernizing garment manufacturing, the study of requirements is the second step, following the definition of value, to save labour hours and improve efficiency through automation. These techniques help identify how automation can enhance the process and reduce the standard time required for garment application. Design requirements define the crucial characteristics that a solution must possess to effectively address the problem in lean thinking. It involves identifying the design requirements to address the reasons for time consumption and waste, including material and process inefficiencies. Automation can significantly improve output precision, reducing the number of failed parts in garment manufacturing. Failures in garment assembly often result in the need for repair work and the scrapping of accessories and fabric. The requirement study in this methodology focuses on estimating the projected improvements in quality pass yield, cycle time reduction, and labour standard time savings. It also emphasizes individual processes such as part sewing, embroidery, stitching, ironing, and packing.

*4.3 Explore modernization solutions and select the best solution*

Numerous effective explorations of jig & fixture, automation, and modernized digital solutions for solving design problems exist. Each potential solution should be evaluated based on how well it meets the design requirements. Some solutions may fulfil more requirements than others, so it's important to reject those not meet the necessary criteria. By comparing factors such as expected output rate, SAM, labour content, and enhanced efficiency percentage among viable solutions, it becomes clear that certain options offer benefits in terms of increased earnings and cost savings. The total investment cost encompasses the purchase of equipment or machines, additional costs of accessories and fabric, and the development cost of building the new automated machine, semi-automated machine, jig and fixture, and training costs if the machine is built in-house. The total investment cost and the earning income are used to calculate the Return on Investment (ROI) for various potential enhancement solutions. The best solution can be selected by weighing the costs and benefits of the enhancement process, with a focus on achieving a high success rate for new modernization solutions.

If the proposed solution cannot meet the requirement successfully, it requires redesigning the automation solution and related machinery in return the build a prototype from the automated machine stage.

*4.4 Develop the best solution*

The development process involves multiple iterations and redesigns of the solution until the final acceptance of test criteria in function, assembly performance, quality, and sustainability by various garment assemblies. The internal in-house testing of the modernization solution is necessary to identify any new problems, make adjustments, and test new solutions before settling on the final best solution. Garment modernization solutions encompass new cutting machines, automated machines, intelligent hanger lines for the stitching process, needle collection and inspection machines, ironing machines, automated packing machines, digitalization equipment for intelligent manufacturing, and other components.

*4.5 Build a Prototype from automated machinery*

A prototype of cutting fabric pieces, part assembly garment and the finished garment from any new automated machine, semi-automated machine or template & fixture are important to prove the garment whether satisfies the style design requirement and quality requirements. In the change of fabric, accessories and thread for lean standardization and automation, it is made with different materials and garment assemble construction, the outlook of garment style and reliability is not same as original style design. A prototype of any changeover whatever automated machine in the manufacturing process, fabric, lining, accessories and garment assembly construction, is required to go through the development quality process and test with customer acceptance.

*4.6 Test and redesign solution*

The design process involves multiple iterations and redesign of the solution until the final acceptance of test criteria in function, assembly performance, quality, and sustainability by various garment assemblies. The internal in-house testing of the modernization solution is necessary to identify any new problems, adjust, and test new solutions before settling on a final solution. Garment modernization solutions include new cutting machines, automated machines, intelligent hanger lines for the stitching process, needle collection and inspection machines, ironing machines, automated packing machines, digitalization equipment for intelligent manufacturing, and other components.

*4.7 Create workflow of the manufacturing process chain*

After reducing waste from the value stream and implementing automation machines, semi-automated machines, and jigs & fixtures, the next crucial step is to ensure uninterrupted and efficient flow. In the garment stitching process, sewing the different parts of the garment—such as the front, back, waist, and hems—is a parallel operation, allowing sewing operators to work on these parts individually before the final garment stitching. Introducing a new workflow involves ensuring a smooth flow of value-adding activities, implementing automation, balancing workloads, establishing cross-functional departments, and training employees to be versatile and adaptable.

*4.8 Map the Value Stream*

Once the solution from the previous steps has been defined, the engineer aims to focus on identifying and mapping the value stream. In [3], lean practices utilize the Value Stream Mapping (VSM) technique to identify the areas that contribute to optimizing the entire manufacturing process and supply chain. Activities that do not contribute to shortened cycle time or enhancing efficiency are considered wasteful. Waste is categorized into two types: non-value added but necessary, and non-value added and unnecessary. It is crucial to eliminate waste while minimizing process optimization. By reducing and eliminating unnecessary processes or steps, automation designers can ensure that customers receive precisely what they need while simultaneously reducing production costs.

If the future state of value stream mapping can be improved the workflow for better performance based on lean thinking and lean practice to eliminate the wastage process or combined process, the planned workflow can be amended to the best solution and return to the previous workflow to stage.

*4.9 Line capacity balancing of new workflow*

The optimization of capacity requirement planning through line balancing is essential for improving the efficiency of job shop scheduling and job order levels. By applying linear programming to the entire production process on a style and time basis, and enhancing its workflow with VSM, we can achieve significant improvements.

Linear Programming for CRP can be utilized in production planning to optimize resource allocation, including labour, machinery, and materials. It aids in determining the most efficient production quantities to meet demand while minimizing costs. However, unexpected events in real manufacturing scenarios, such as machine breakdowns, material shortages, and failures in garment sewing assembly from upstream workstations, often lead to extensions in the stitching standard times of tasks. To achieve stable manufacturing, it would be beneficial to specify the processing period of each work step and operation, forecast the frequency of such unexpected events, and account for delayed operation times in the CRP.

If the result of capacity requirement planning can be improved by the workflow enhancement and production line balance based on the capacity plan to adjust the capacity of machines and the number of machines to maximize the final garment output. The planned workflow can be amended to the best solution and returned to the previous workflow stage. If the machinery is required to be modified to increase the machine capacity, it is required to go back to the prototype from the automation machine stage.

*4.10 Establish Just-in-Time Pull System*

In reference [3], inventory is considered one of the biggest wastes in any production system. The goal of a pull-based system is to limit garment inventory and garment assembly items while ensuring that the necessary materials and information are available for a smooth flow of work. In other words, a pull-based system allows for just-in-time delivery and manufacturing, where products are created at the time they are needed and in the exact quantities needed. Pull-based systems are always designed based on the needs of the end customers. By following the value stream and working backwards through the production system, you can ensure that the products produced will be able to satisfy the needs of customers.

*4.11 Pursue Perfection in automation development*

Referring [5] to the previous section, the pursuit of perfection in lean practice and lean production is crucial for implementing garment modernization. The planning phase has encompassed steps such as Value Stream Mapping (VSM), capacity planning, expected machine scheduling, and lean planning to minimize waste and reduce standard production time for garment modernization, whether through automated machines, semi-automated machines, or new process flows.

In the pursuit of perfection in automation development for garment manufacturing modernization, it is recommended that the lean implementation project committee, led by department heads, oversee the lean modernization planning. Every employee should strive for the perfection of lean implementation. The lean project involves senior management, functional department heads, production planners, facility heads, engineering teams, and production teams. Regular meetings are held by the lean project committee to drive continuous lean projects, ensuring that the lean concept, plan, action, and review are in place to propose the next lean proposal and plan as part of an ongoing cycle.

A common problem in garment workpieces is them getting jammed in bottleneck processes and machines. Line balancing can be used for robust capacity planning and to identify process bottlenecks, not solely relying on the isolated implementation of automated machines. Applying automated machines for specific processes with capacity planning and machine scheduling techniques can resolve production line capacity imbalance issues.

The lean methodology has significant benefits for garment modernization, as it can increase garment production output and efficiency by addressing capacity planning and minimizing process and production waste.

## V. LEAN METHODOLOGY IMPLEMENTATION IN GARMENT MANUFACTURING

In China, garment factories have implemented lean methodology for modernizing garment manufacturing and implementing automation. The application of lean principles to garment manufacturing workflow and automation development can be seen across many major corporations, including garment factories.

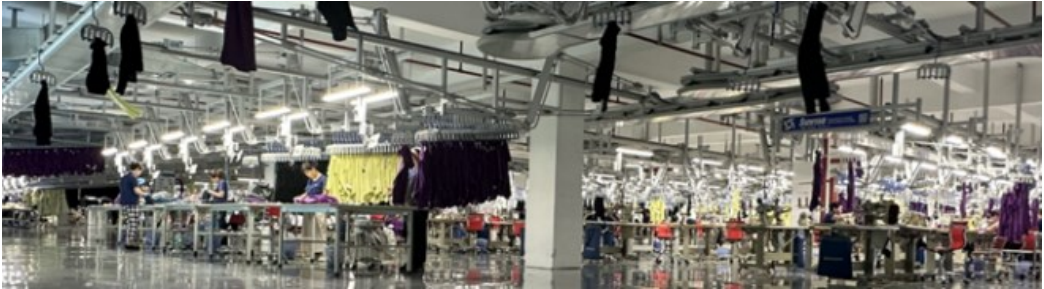

Figure 2 Intelligent Hanger Line and Hanger System from Sunrise Company

Company X and its factory in the southern region of China have applied lean methodologies to eliminate redundancies and reduce the work hours necessary to complete tasks. Previously, the factory had run its day-to-day operations in traditional garment sewing without implementing a hanger line and hanger system to improve its efficiency, as shown in Figure 2.

*5.1 Define value and requirement study*

The engineering team proposed the proposal of intelligent hanger how to improve the movement of the part sewing assembles within the production line. The production line uses the conveyor belt line to carry the batch of part assembly of garments. The proposal of an intelligent hanger line for garments has shown more benefits how in improving efficiency and shortening the transportation time between sewing stations.

*5.2 Explore automation to the sewing operation*

The requirement has been defined to apply the hanger system and hanger line to replace the manual flow. The one-piece workflow is required to resolve the batch flow in the sewing line. The supplier hanger system and hanger line can satisfy the mentioned requirement.

The benefits of a hanger system and hanger line can improve by over 5%-10% efficiency from supplier experience based on the supplier assumption of the number of part movement frequency such as how many pieces of part sewing piece pass to cellular sewing line and move time per bundle batch. The hanger line is focused on eliminating the move time of part sewing pieces. The estimation of cycle time is shortened to implement the hanger line. Following the methodology, the lean principle can be applied to define which operation and work steps in the process are being eliminated to reduce the non-value operations and steps.

*5.3 Select the best solution of automation for the sewing operation*

To estimate the potential value, we have investigated past results and observed a 5%-15% improvement in efficiency from our hanger supplier. After conducting a thorough analysis of our sewing lines, we have identified opportunities to implement new automation and advanced manufacturing techniques. Our goal is to optimize processes, reduce standard cycle times, and minimize the reliance on manual labour, all while adhering to the principles of one-piece flow in our hanger line.

Company X has meticulously evaluated various hanger system suppliers to find the best fit for our garment modernization initiative. After comparing factors such as functionality, features, pricing, and service, Sunrise has been selected as the supplier of choice. Sunrise offers an intelligent hanger system and line that is well-suited for our garment manufacturing needs, including large daily output, multi-order, multi-style, and multi-size production. Their cross-regional multi-layer and zoning intelligent hanging sorting system, integrated with quality inspection, ironing, and packaging processes, effectively streamlines garment transportation and handling. This results in standardized, error-free production and significant cost savings in labour, operations, and space, ultimately leading to improved efficiency.

*5.4 Develop, build and test the hanger line as the best solution*

Factory X has implemented the intelligent suspension production system models S160 and S180 to meet its production needs effectively. The system chain and track feature a patented design to ensure stable mechanical operation, as well as safety and cleanliness in the environment. Figure 3 shows a digital barcode scanner that records the bundle ticket in the hanger line, eliminating the need for manual recording of production output by bundle ticket. The hanger line and its digital system can address the issue of in-line stock, thereby reducing the overall order completion lead time.

The optimal solution was provided by the supplier, Sunrise, rather than using in-house machinery. Factory X's project team invested time in visiting the supplier's demonstration and providing garment assembly samples to the supplier workshop for solution development. The supplier incorporated a tailor-made feature in the hanger line to meet the specific requirements of garment assembly.

The supplier conducted the building and testing of the tailor-made feature hanger systems at their site. Any issues were addressed and resolved by the supplier in their workshop. Upon successful testing, the entire hanger system, hanger line, and solution were migrated to Factory X. This process is a typical implementation approach to ensure project success within a short time frame.

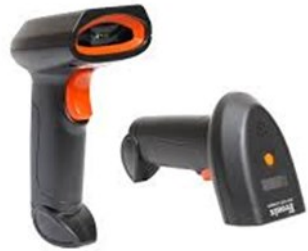

Figure 3 Bar code scanner and its record System

*5.5 Map the Value Stream*

Value Stream Mapping (VSM) is used to identify and visualize value-added and non-value-added activities. VSM is very effective for improvement in lead time and helps to reduce the overall cost of production in company X according to Mahadeo M. Narke, C. T. Jayadeva [11]. In this case, the purpose may not be entirely clear before the analysis is done, which leads us to add all the known data about the process. The list of critical information below gives an overview of process data and abbreviations that may be used for a VSM and capacity requirement plan.

- Customers demand quantity per style in pieces
- Cycle time (C/T) in hours or minutes
- Added Value in hours or minutes
- Non-Added Value in hours or minutes
- Changeover time (C/O) in minutes
- Number of operators (Op) in person
- Available time in minutes
- Uptime/downtime in minutes
- Quality or defects rate (Q) in the number of parts of the garment pieces
- Inventory levels in pieces between operations

The purpose of the Value Stream Mapping Diagram in this case study is to maximize added value and minimize non-value operations and waste in various garment styles. The current state of VSM can identify the major long lead time operations. The cycle time per style in the high mix low volume production is based on the time study from an industrial engineer. The Current State VSM is meant to capture the "as is" situation. Current State VSM is neither good nor bad; it simply shows the current figures and VSM diagram in Figure 4 before lean execution and automation execution. In other words, the team must document exactly what we see and what we find.

In the envisioned future state of VSM, the garment manufacturing process is poised to employ the most effective solution to meet the operational requirements outlined in the lean methodology diagram in Figure 1. In this scenario, the intelligent hanger system and hanger line are integrated into the primary sewing operation, utilizing a one-piece workflow. The workflow is optimized for batch layout and cellular production line integration with the hanger line.

Despite the initial challenges associated with the implementation of the new hanger line and hanger system, the benefits far outweigh the setup difficulties. Any member of the operations team can scan bundle information to track real-time output, and rerun reports at any time to access the most current data available. By streamlining processes and saving time, both in recordkeeping and operations visibility, this approach adds significant value to customers. Real-time status updates can be promptly provided to any customer inquiry by any operations team member.

The current lead time in the Value Stream Mapping (VSM) stands at 16,876 minutes. The integration of the intelligent hanger system and hanger line with the main sewing process and part sewing process has resulted in a reduction of the sewing operation's cycle time. This improvement is attributed to the hanger system's ability to minimize move time and upload/unload time. Additionally, the uptime percentage has increased due to the hanger line, and the system's digitalization capabilities have led to reduced downtime by capturing output automatically. Looking ahead, the future state of VSM is projected to achieve a lead time reduction of 8,896 minutes, leveraging the benefits of automation to align with the lean methodology for garment modernization.

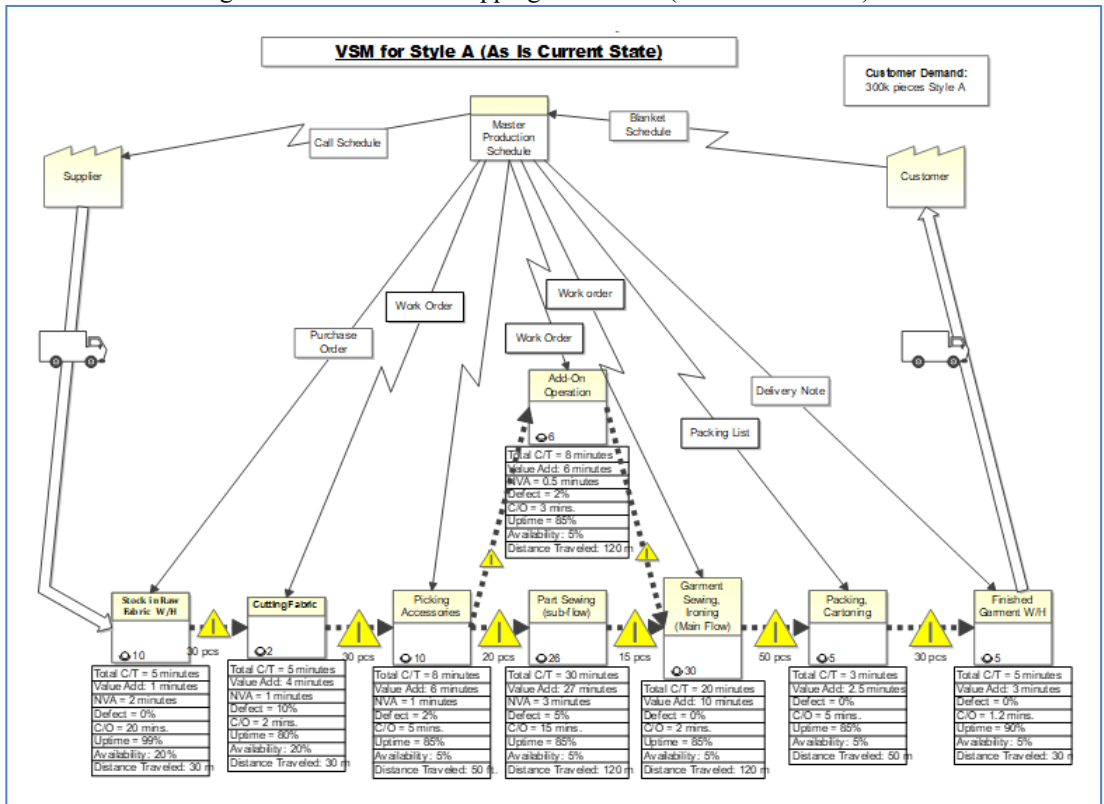
Figure 4 Value Stream Mapping Worksheet (as is current state)

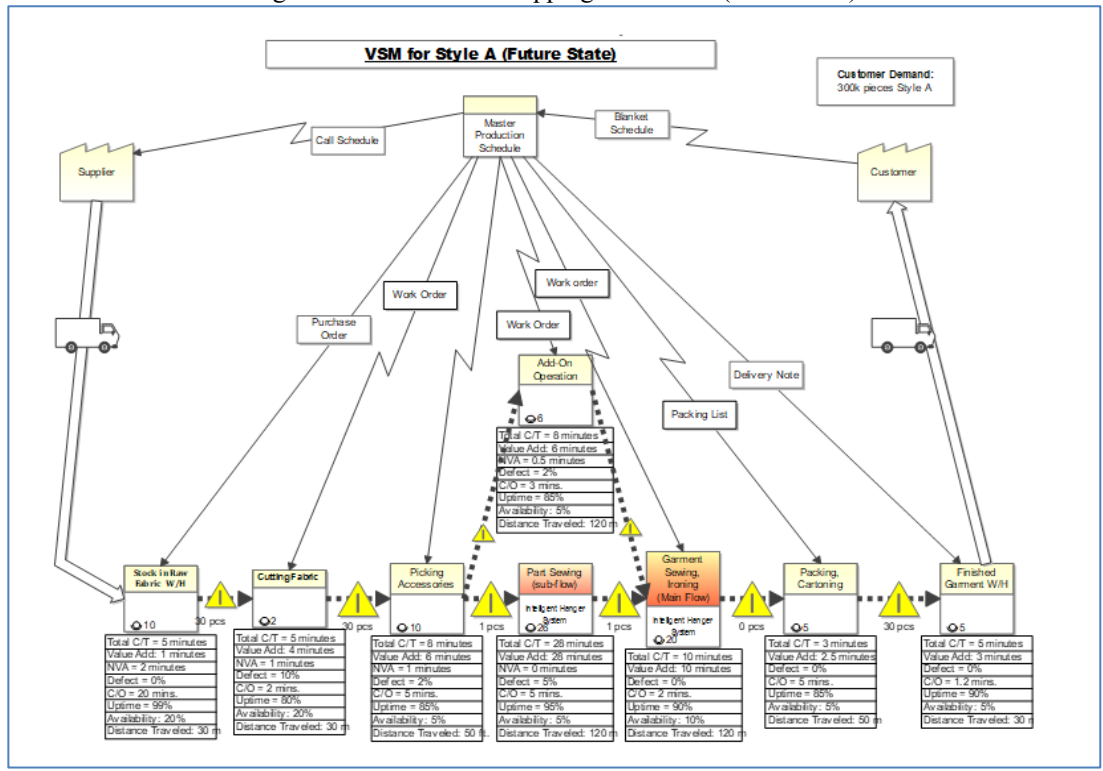
Figure 5 Value Stream Mapping Worksheet (future state)

In this context, the garment factory's operations encompass not only the production of singular styles but also the manufacturing of various styles at high mixed and low volumes. The VSM's ability to utilize the weight average method in calculating cycle time based on style combinations limits its accuracy in depicting the current state of VSM and projecting the future VSM.

*5.6 Build a new workflow and work for capacity requirement planning*

The new workflow is created for manufacturing modernization and automation according to the lean methodology. The capacity requirement planning can follow the new workflow to calculate the hanger system and automated machines whether provide the sufficient capacity of machine and work center to support for garment manufacture.

Capacity requirement planning is the next step after VSM. CRP method from Daisuke Morita adopts the mathematical method to calculate the machine capacity as a constraint to make a line balance to enhance the production output and efficiency.

The time period $t = 1,2,...T$ during which operation orders are to be processed. The processing period of each operation order is determined so that the total resource requirements for each manufacturing resource incurred by operation orders at every unit time never exceeds the predetermined capacity. Let $W = \{1,....,c\}$ denote a set of work centers in the manufacturing system. Work center $k$ ($k = 1,...,c$), consists of $N_k^W$ identical parallel machines; $M_{k1},...M_{kN_k^W}$. Production capacity over a finite planning horizon, $C_k$, at work center k can be expressed by

$$C_k = c_k N_k^W \quad (1)$$

where $c_k$ denotes the maximum production capacity per machine in work center k.

According to the simplex method (Halaç, 1995:363-410) in linear programming from Daisuke Morita [7], firstly the mathematical model is established and then the variables are added to the inequalities in the model and an equality is obtained and formulated. After that, objective function and constraints are placed in the simplex table of Constraints show that the labour force, machines and other resources held by the decision-makers are limited.

There are several factors to be taken into consideration in the theoretical structure of linear programming: objective function, constraints and positivity condition.

These are
$x_j$: Decision variables (like production or cost amounts),
$c_j$: Unit profit or cost coefficient,
$b_i$: Resource capacity,
$a_{ij}$: Technical coefficient

$$Z = \sum_{j=1}^{n} c_j x_j \quad (j = 1,2,...,n) \quad (2)$$
$$\sum_{j=1}^{n} a_{ij} x_j \leq b_j \quad (i = 1,2,...,m) \quad (3)$$
$$x_j \geq 0 \; (j = 1,2,...,n) \quad (4)$$

To apply the maximize the profit, the formula is shown below:

$$Z_{max} = \sum_{j=1}^{n} c_j x_j \quad (j = 1,2,...,n) \quad (5)$$

Where $Z_{max}$ is the maximize the profit.

$$\sum_{j=1}^{n} a_{ij} x_j \leq b_j \quad (6)$$

Where $x_j$ is the production quantity, x in the operation j, $a_{ij}$ is the standard applied minutes of style in the operation j, $b_j$ is the machinery capacity in operation j in the daily plan or hourly plan.

The developed computational model has been analyzed to obtain optimal solution values. In dealing with machinery capacity constraints, it may be beneficial to increase capacity by adding more machines, semi-automated machines, and jigs & fixtures, thereby shifting the constraint from labour to machinery in the automation development of garment modernization.

To fulfil capacity requirements and review the demand of machines and suppliers for various seasonal styles, cutting fabric machines, sewing machines, embroidery machines, pocket welting machines, template sewing machines, and special add-on process machines are necessary.

Company X is fully committed to modernizing its garment manufacturing processes. The project team has been established, and they are learning to apply lean thinking and concepts using lean methodology for garment manufacturing modernization, including automation. The company has successfully revamped many important processes and gained valuable experience in implementing lean, automation, and modernization. According to the final step of the lean methodology for garment modernization, continuous driving actions are essential to sustain the loop of improvement. As high technology continues to advance in the garment industry, the trend can persist in the right direction.

## VI. RESULT

When comparing the traditional implementation methodology with the lean methodology for garment modernization, it's evident that the traditional methodology does not take into account any capacity requirement planning (CRP) in the plan. The traditional approach focuses on increasing output by enhancing automation and the hanger system at isolated work centers to shorten the cycle time of sewing operations. According to the capacity analysis of the traditional methodology shown in Figure 6, the projected output of the future state of value stream mapping (VSM) is expected to increase from 200pcs/day to 240pcs/day after implementing automation. However, the designer and planner expect a garment output of 300pcs/day from the operation without reviewing the operation bottleneck.

On the other hand, the lean methodology for garment modernization combines lean principles with capacity requirement planning to address capacity constraints under the theory of constraints. By utilizing capacity requirement planning, the designer and planner can predict the expected output of each operation and work center, as well as identify and address the bottleneck operation. In garment manufacturing, labour-intensive operations can increase labour to achieve line balance, and in the case of fabric-cutting operations, manual cutting can be increased from 1 team to 3 teams to boost output until the bottleneck operation shifts back to a critical capacity constraint, such as a machine-intensive operation as shown in Figure 7. Increasing the number of machines to boost capacity results in a corresponding increase in final garment output. The lean methodology emphasizes the continuous review of non-value-added processes to explore automation solutions for further improvements.

Figure 6 Capacity analysis of the traditional methodology

Figure 7 Capacity analysis of the lead methodology with CRP in the automated machine implementation

## VII. CONCLUSION

**Traditional Methodology**
Theoretical Current State (before applied automation to reduce the cycle time in sewing operation)

| Work Station | Process | (a) Batch Qty. kit unit | (b) Total Time min/batch | (c) =(b)/(a) Cycle Time min/kit unit | (d) Operators | (e) = (d)x(360/c) Daily Output (10hr) kit unit/day | (f) Machine Daily Capacity kit/day | (g) No. of Machine set | (h)=(f) x (g) Total Machine Daily Capacity kit/day | |
|---|---|---|---|---|---|---|---|---|---|---|
| 1 | Fabric W/H Receiving | 1000 | 200 | 5 | 3 Teams | 360 | | | N/A | |
| 2 | Fabric Cutting | 1000 | 200 | 5 | 2 Teams | 240 | 120 | 2 | 240 | |
| 3 | Picking Accessories | 1440 | 180 | 8 | 4 Teams | 300 | | | N/A | |
| 4 | Part Sewing | 1 | 30 | 30 | 10 | 200 | 20 | 10 | 200 | (Bottleneck) |
| 5 | Add-On Processes (Embroidery, Pocket Welting, Template Sewing,,, etc) | 1 | 8 | 8 | 5 | 375 | 75 | 5 | 375 | |
| 6 | Finished Garment Sewing | 1 | 20 | 20 | 10 | 300 | 30 | 10 | 300 | |
| 7 | Packing, Cartoning | 1 | 3 | 3 | 3 Teams | 600 | | | N/A | |
| 8 | Finished Garment W/H Delivery | 400 | 80 | 5 | 3 Teams | 360 | | | N/A | |
| | | | | | FG Output (pcs) : | 200 | | FG Machinery Output (pcs): | 200 | |

Theoretical Future State (after applied automation to reduce the cycle time in sewing operation without run CRP)
Not run CRP that cannot find out the bottleneck operation.

| Work Station | Process | (a) Batch Qty. kit unit | (b) Total Time min/batch | (c) =(b)/(a) Cycle Time min/kit unit | (d) Operators | (e) = (d)x(360/c) Daily Output (10hr) kit unit/day | (f) Machine Daily Capacity kit/day | (g) No. of Machine set | (h)=(f) x (g) Total Machine Daily Capacity kit/day | |
|---|---|---|---|---|---|---|---|---|---|---|
| 1 | Fabric W/H Receiving | 1000 | 200 | 5 | 3 Teams | 360 | | | N/A | |
| 2 | Fabric Cutting | 1000 | 200 | 5 | 2 Teams | 240 | 120 | 2 | 240 | (Bottleneck) |
| 3 | Picking Accessories | 1440 | 180 | 8 | 4 Teams | 300 | | | N/A | |
| 4 | Part Sewing | 1 | 20 | 20 | 10 | 300 | 30 | 10 | 300 | Reduced C/T |
| 5 | Add-On Processes (Embroidery, Pocket Welting, Template Sewing,,, etc) | 1 | 4 | 4 | 5 | 750 | 150 | 5 | 750 | Reduced C/T |
| 6 | Finished Garment Sewing | 1 | 10 | 10 | 10 | 600 | 60 | 10 | 600 | Reduced C/T |
| 7 | Packing, Cartoning | 1 | 3 | 3 | 3 Teams | 600 | | | N/A | |
| 8 | Finished Garment W/H Delivery | 400 | 80 | 5 | 3 Teams | 360 | | | N/A | |
| | | | | | FG Output (pcs) : | 240 | | FG Machinery Output (pcs): | 240 | |

Comment: Because there is no run CRP, although the cycle time in part sewing, add-on process and the finished garment has been reduced, the final garment is only increased to 240pcs.day, not 300pcs/day.

*Annotations:* The bottleneck operaton is shifted to from part sewng to fabric cutting operation. Designer expects to apply the automation for shortening the cycle time in sewing related operations as the minum output, 300pcs/day.

The article introduces a new lean methodology for modernizing garment manufacturing, focusing on lean thinking, lean practice, automation development, VSM, and CRP, and how to integrate them effectively. The implementation of isolated automated machines

**Lean Methodology (lean with CRP)**
Theoretical Future State (after applying automation to reduce the cycle time in sewing operation with CRP)
Not run CRP to find out the bottleneck operation.

| Work Station | Process | (a) Batch Qty. kit unit | (b) Total Time min/batch | (c) =(b)/(a) Cycle Time min/kit unit | (d) Operators | (e) = (d)x(360/c) Daily Output (10hr) kit unit/day | (f) Machine Daily Capacity kit/day | (g) No. of Machine set | (h)=(f) x (g) Total Machine Daily Capacity kit/day | |
|---|---|---|---|---|---|---|---|---|---|---|
| 1 | Fabric W/H Receiving | 1000 | 200 | 5 | 3 Teams | 360 | | | N/A | |
| 2 | Fabric Cutting | 1000 | 200 | 5 | 3 Teams | 360 | 120 | 3 | 360 | |
| 3 | Picking Accessories | 1440 | 180 | 8 | 4 Teams | 300 | | | N/A | |
| 4 | Part Sewing | 1 | 20 | 20 | 10 | 300 | 30 | 10 | 300 | (Bottleneck) |
| 5 | Add-On Processes (Embroidery, Pocket Welting, Template Sewing,,, etc) | 1 | 4 | 4 | 5 | 750 | 150 | 5 | 750 | Reduced C/T |
| 6 | Finished Garment Sewing | 1 | 10 | 10 | 10 | 600 | 60 | 10 | 600 | Reduced C/T |
| 7 | Packing, Cartoning | 1 | 3 | 3 | 3 Teams | 600 | | | N/A | |
| 8 | Finished Garment W/H Delivery | 400 | 80 | 5 | 3 Teams | 360 | | | N/A | |
| | | | | | FG Output (pcs) : | 300 | | FG Machinery Output (pcs): | 300 | |

Comment: Because there is no run CRP, although the cycle time in part sewing, add-on process and the finished garment has been reduced, the final garment

*Annotations:* After run CRP, desiger knows the bottleneck operaton and then increase the machine and team to fabric cutting operation, so the bottleneck operaion has been shifted to the part sewing operaiton s constraint of machine. The bottleneck operation is the part sewing operation because of constraints of capacity. The bottleneck output is 300pcs/day.

has shown improvement in efficiency and cycle time reduction for specific production workstations. However, the overall garment output and efficiency have not been significantly enhanced due to the narrow focus on automation. To address this, a holistic approach considering the entire production line and process is proposed using VSM and CRP to achieve production line and production center balance, leading to improved efficiency, reduced manufacturing costs, labour time, and lead time, thereby adding value to the company and factory.

Additionally, the article presents a practical example of utilizing these tools to overcome challenges and offers a continuous improvement solution for extending the lean methodology in garment manufacturing modernization.

## VIII. ACKNOWLEDGMENT

Prof. Kong Wai Man, Ray wishes to thank our assistance and engineers of from Eagle Nice (International) Holding Ltd.

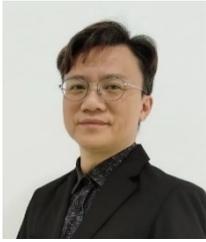

**Ray Wai Man Kong** (Senior Member, IEEE) Hong Kong, China. He received a Bachelor of General Study degree from the Open University of Hong Kong, Hong Kong in 1995. He received an MSc degree in Automation Systems and Engineering and an Engineering Doctorate from the City University of Hong Kong, Hong Kong in 1998 and 2008 respectively.

From 2005 to 2013, he was the operations director with Automated Manufacturing Limited, Hong Kong. From 2020 to 2021, he was the Chief Operating Officer (COO) with Wah Ming Optical Manufactory Ltd, Hong Kong. He is currently a modernization director with Eagle Nice (International) Holdings Limited, Hong Kong. He holds an appointment, as an Adjunct Professor of the System Engineering Department at the City University of Hong Kong, Hong Kong. He published an Incremental Model-based Test Suite Reduction with Formal Concept Analysis in the Journal of Information Processing Systems in June 2010. ISSN: 2092-805X in Korea Information Processing Society. His research interest focuses on Intelligent Manufacturing, Automation, Maglev Technology, robotics, Mechanical Engineering, Electronics and System Engineering for industrial factories.

Prof. Dr. Kong Wai Man, Ray is Vice President of CityU Engineering Doctorate Society, Hong Kong and a Consulting Member of the Doctors Think Tank Academy, Hong Kong. He has published 5 intellectual properties and patents in China.

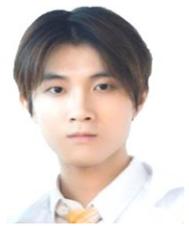

**Theodore Ho Tin Kong** (MIEAust, Engineers Australia) received his Bachelor of Engineering (Honours) in mechanical and aerospace engineering from The University of Adelaide, Australia, in 2018. He then earned a Master of Science in aeronautical engineering (mechanical) from HKUST - Hong Kong University of Science and Technology, Hong Kong, in 2019.
He began his career as a Thermal (Mechanical) Engineer at ASM Pacific Technology Limited in Hong Kong, where he worked from 2019 to 2022. Currently, he is a Thermal-Acoustic (Mechanical) Design Engineer at Intel Corporation in Toronto, Canada. His research interests include mechanical design, thermal management and heat transfer, and acoustic and flow performance optimization. He is proficient in FEA, CFD, thermal simulation, and analysis, and has experience in designing machines from module to heavy mechanical level design.

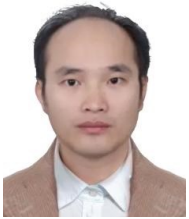

**Tianxu Huang**, BEng, 1984, Guangxi City Vocational University, 1 Luoyue Avenue, Jiangzhou District, Chongzuo City, Guangxi, China.
He holds the position of senior experimentalist at the School of Information Engineering, Guangxi City Vocational University, specializing in big data engineering and computer application engineering.
He has made significant contributions to the field through his extensive research and publications. His work includes numerous research papers and textbooks widely recognized in the academic community.